\documentclass[10pt,twocolumn,letterpaper]{article}

\usepackage{cvpr}              

\usepackage{algorithm}
\usepackage{algorithmic}
\usepackage{listings}
\usepackage{framed}
\usepackage{bbold}
\usepackage{colortbl}
\usepackage{kotex} 
\newcommand{\figref}[1]{Fig.~\ref{#1}}

\newcommand{\tabref}[1]{Table~\ref{#1}}
\newcommand{\eqnref}[1]{Eq.~(\ref{#1})}

\newcommand{\Figref}[1]{Figure~\ref{#1}}

\newcommand{\ourmodel}{GaussianMotion}

\definecolor{cvprblue}{rgb}{0.21,0.49,0.74}
\definecolor{cellcol}{gray}{.92}
\usepackage[pagebackref,breaklinks,colorlinks,allcolors=cvprblue]{hyperref}

\def\paperID{9343} 
\def\confName{CVPR}
\def\confYear{2025}

\title{GaussianMotion: End-to-End Learning of Animatable \\ Gaussian Avatars with Pose Guidance from Text}

\author{Gyumin Shim\\
Korea Advanced Institute of Science and Technology\\
{\tt\small shimgyumin@kaist.ac.kr}
\and
Sangmin Lee\\
Sungkyunkwan University \\
{\tt\small sangmin.lee@skku.edu}
\and
Jaegul Choo\\
Korea Advanced Institute of Science and Technology\\
{\tt\small  jchoo@kaist.ac.kr}
}

\begin{document}
\twocolumn[{%
\renewcommand\twocolumn[1][]{#1}%
\maketitle
\centering
\includegraphics[width=1\linewidth]{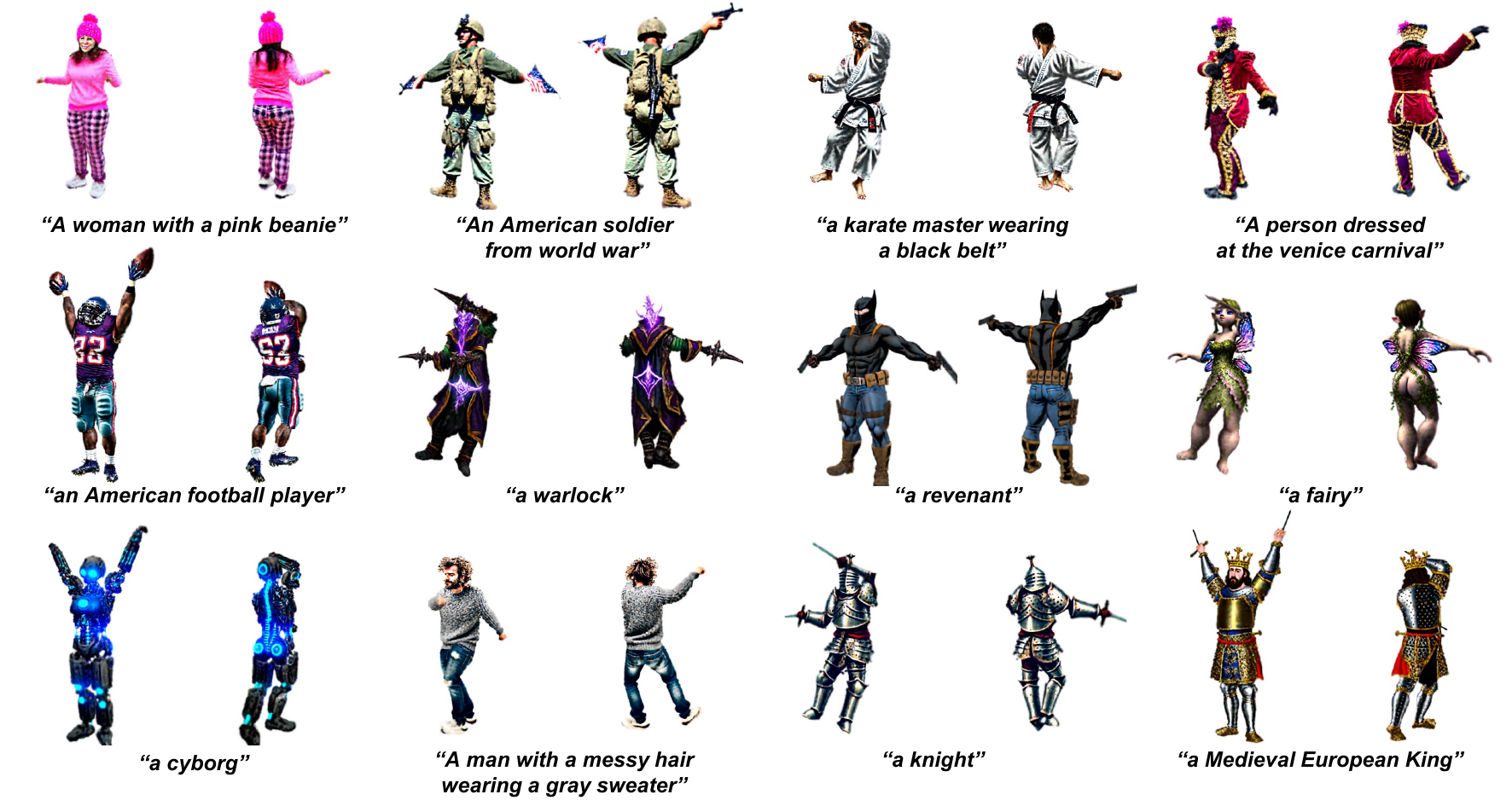}
\captionof{figure}{{\bf Examples of 3D human models generated by \ourmodel.}
Our method is able to generate high-quality Gaussian-based avatars from text and render animated scenes from user-specified pose inputs.
}
\label{fig:teaser}
}]

\begin{abstract}

In this paper, we introduce \ourmodel, a novel human rendering model that generates fully animatable scenes aligned with textual descriptions using Gaussian Splatting.
Although existing methods achieve reasonable text-to-3D generation of human bodies using various 3D representations, they often face limitations in fidelity and efficiency, or primarily focus on static models with limited pose control.
In contrast, our method generates fully animatable 3D avatars by combining deformable 3D Gaussian Splatting with text-to-3D score distillation, achieving high fidelity and efficient rendering for arbitrary poses.
By densely generating diverse random poses during optimization, our deformable 3D human model learns to capture a wide range of natural motions distilled from a pose-conditioned diffusion model in an end-to-end manner.
Furthermore, we propose Adaptive Score Distillation that effectively balances realistic detail and smoothness to achieve optimal 3D results.
Experimental results demonstrate that our approach outperforms existing baselines by producing high-quality textures in both static and animated results, and by generating diverse 3D human models from various textual inputs.


\end{abstract}

\section{Introduction}

The demand for reconstructing and rendering 3D avatars has surged with advancements in computer graphics, enabling a wide range of applications, including virtual reality and metaverse content.
Building on diverse 3D representations~\cite{wang2021neus, shen2021deep, mildenhall2021nerf, kerbl20233d}, numerous studies have explored the reconstruction of 3D human avatars from various data sources, including 3D scans~\cite{saito2019pifu, saito2020pifuhd, zheng2021pamir}, video sequences~\cite{weng2022humannerf, peng2021animatable, hu2024gauhuman}, single images~\cite{huang2022one, huang2024tech}, and even text~\cite{kolotouros2024dreamhuman, cao2024dreamavatar, liao2024tada, liu2024humangaussian}.
In particular, generating 3D human models from text is significantly challenging, as textual descriptions provide limited information about the 3D structure of the human body, which has complex articulations.

As text-to-image diffusion models~\cite{rombach2022high, saharia2022photorealistic, podell2023sdxl} have emerged, capable of synthesizing diverse and realistic images from textual information, their applicability has significantly expanded across various research fields.
The introduction of score distillation by DreamFusion~\cite{poole2022dreamfusion} marked a breakthrough, enabling text-to-3D synthesis using 2D diffusion models without requiring labeled 3D data. 
Building upon this technique, subsequent studies~\cite{wang2024prolificdreamer,tang2023dreamgaussian,yi2023gaussiandreamer,lin2023magic3d} have further explored the generation of detailed 3D models from text in an unsupervised manner. 

The text-to-3D task includes the generation of 3D {\it human} models, demonstrating the potential of this approach to create realistic and detailed representations of human models based on textual descriptions.
Recently, several approaches leveraging neural representations~\cite{mildenhall2021nerf} or mesh representations~\cite{shen2021deep} have been proposed~\cite{kolotouros2024dreamhuman, cao2024dreamavatar, liao2024tada, huang2024dreamwaltz, huang2024humannorm}. 
However, these methods often face challenges in balancing both fidelity and efficiency. Neural representations typically incur high computational costs due to the large number of point samples along each ray, particularly when rendering high-resolution images, while mesh-based methods struggle to preserve fine-grained details due to limitations in resolution and structure.

With the emergence of 3D Gaussian Splatting~\cite{kerbl20233d}, a powerful 3D model that surpasses neural representations in terms of efficiency, several Gaussian-based human rendering methods have also emerged.
Yuan et al.\cite{yuan2024gavatar} proposed a hybrid method that combines mesh and Gaussian representations, using Signed Distance Functions (SDF)~\cite{park2019deepsdf} to regularize the opacity of the Gaussian.
HumanGaussian~\cite{liu2024humangaussian} introduced a fully Gaussian-based approach, demonstrating competitive generation quality and enabling efficient rendering of full-body human images during training and inference.
However, HumanGaussian focuses on static models, which limits its applicability in dynamic scenarios where handling complex pose variations and animations is critical.




To address the limitations of fidelity, efficiency, and pose variation, we propose \ourmodel, a novel human rendering model that generates realistic and animatable scenes from textual descriptions. Our approach enables end-to-end training of animatable Gaussian avatars without relying on regularization from other representations, capturing fine-grained geometric details purely through the strengths of Gaussian Splatting.

\begin{figure*}[t]
\centering
\begin{tabular}{@{}c}
\includegraphics[width=1\linewidth]{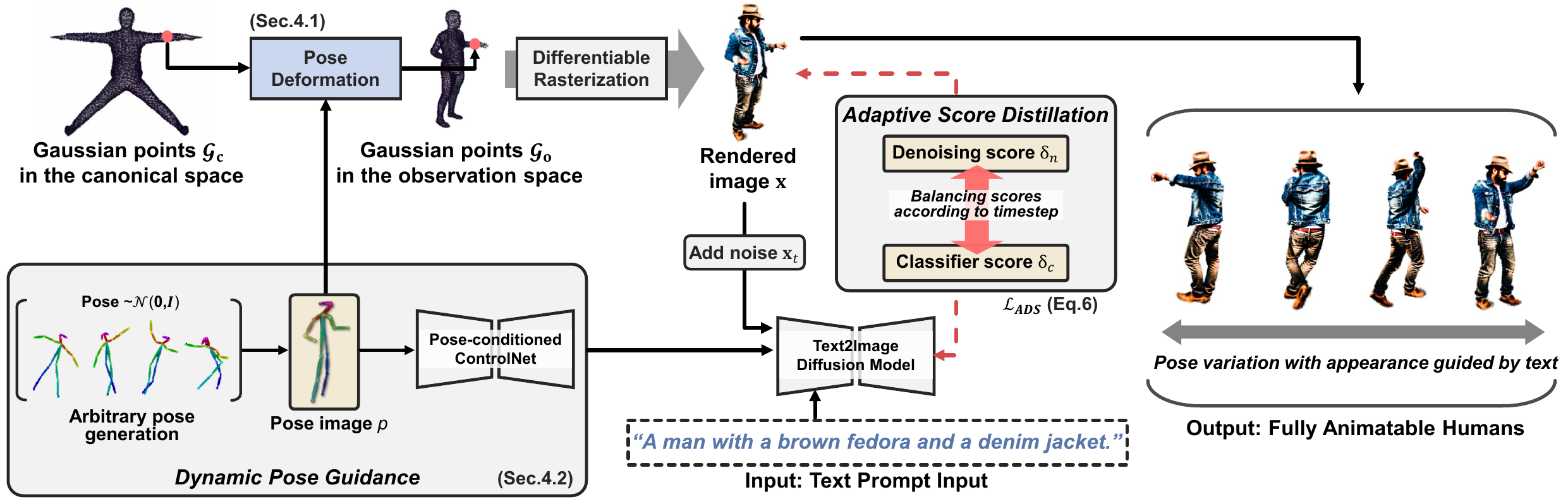}
\end{tabular}
\caption{{\bf Overview of our proposed framework.}
Given a text prompt as input, we generate animatable 3D humans by modeling deformable Gaussian Splatting, where Gaussian points adapt their positions based on input poses. 
The points are defined in a canonical space and shared across different poses (observation spaces). Random poses are sampled to deform the Gaussian points and rendered as pose images to provide pose-aware guidance for the rendered images $\mathbf{x}$ through score distillation. 
After optimizing the Gaussian points to reflect the appearances described by the text prompt, fully animatable scenes are rendered based on user-specified input poses during inference.
}
\label{fig:framework}
\end{figure*}

As depicted in \Figref{fig:framework}, our method integrates deformable 3D human Gaussian Splatting with pose-aware text-to-3D score distillation.
Gaussian points in the canonical space are optimized to capture both the appearance aligned with the given text and pose articulation as the input pose changes.
A key innovation of our model is the use of densely generated random poses with explicit pose guidance during optimization. 
This allows the deformable 3D human model to effectively learn a diverse range of poses distilled from a pre-trained pose-conditioned diffusion model in an end-to-end manner, providing robust pose-aware guidance when rendering complex poses during training. 
Additionally, to achieve optimal results with realistic details, we propose Adaptive Score Distillation as an alternative to naive score distillation sampling (SDS)~\cite{poole2022dreamfusion},
which balances the preservation of fine details while minimizing undesired noise, effectively handling the trade-off between smoothness and high uncertainty.


As a result, our method enables the creation of diverse 3D human models based on the provided text, capturing intricate texture details and supporting realistic animations according to user-specified input poses, all while maintaining computational efficiency. We validate our approach through extensive experiments, demonstrating that it significantly outperforms existing baselines.

\noindent\ In summary, our contributions are as follows:
\begin{itemize}
\setlength\itemsep{-0em}
    \item A novel human rendering model with deformable Gaussian Splatting to create 3D human models aligned with textual descriptions, capable of exhibiting a wide variety of motions.
    \item An innovative framework that optimizes Gaussian points by generating random poses during training, allowing the model to learn both detailed appearances from text descriptions and complex pose articulations through explicit pose guidance.
    \item Adaptive Score Distillation, an improved strategy over naive SDS, effectively balancing the issues of over-saturation and high variance to achieve optimal results with realistic details.
\end{itemize}


\section{Related Works}
\subsection{3D Gaussian Avatar}

Recently, with the emergence of 3D Gaussian Splatting~\cite{kerbl20233d}, which has demonstrated powerful performance in various 3D applications, numerous studies in 3D avatar modeling have increasingly leveraged this technique to create high-quality human models. A range of methods~\cite{hu2024gauhuman, hu2024gaussianavatar, lei2024gart, qian20233dgs, zielonka2023drivable} proposes innovative approaches for reconstructing human avatars using Gaussian representations that can respond dynamically to various movements and expressions. 
These studies draw inspiration from human deformation concepts derived from deformable neural representations~\cite{gao2023neural, peng2021neural, peng2021animatable, weng2022humannerf}, which address how 3D coordinates on a human model are deformed across different poses.
Furthermore, more sophisticated forms of 3D human avatars have been developed, such as ExAvatar~\cite{moon2024expressive}, which incorporates facial and hand expressions, and UV Gaussian~\cite{jiang2024uv}, a hybrid form of animatable avatar that jointly learns mesh deformation and 2D Gaussian textures.
After reconstructing an avatar from monocular or calibrated multi-view videos, these methods facilitate the rendering of scenes from arbitrary viewpoints and poses using the trained 3D Gaussian points during inference, leveraging the computational efficiency of Gaussian representations. 
In this work, we introduce a novel method that can produce an animatable Gaussian avatar from text without requiring any image ground truths.


\subsection{Text-to-3D Human Generation}

Text-to-3D is a popular task which is to generate 3D models from input textual descriptions without relying on text-3D paired data.
Early work utilizing CLIP~\cite{radford2021learning} embeddings to optimize 3D shapes~\cite{sanghi2022clip} or neural representations~\cite{jain2022zero} has successfully demonstrated that 3D objects can be generated solely from textual descriptions.
As DreamFusion~\cite{poole2022dreamfusion} introduces a method to distill priors from pre-trained 2D diffusion models for targeting 3D models, numerous text-to-3D methods~\cite{tang2023dreamgaussian, yi2023gaussiandreamer, wang2024prolificdreamer} have emerged, aiming to generate high-quality 3D models from input textual descriptions by leveraging various diffusion models.
These methodologies can be directly extended to the task of generating 3D {\it humans}, with DreamHuman~\cite{kolotouros2024dreamhuman} and DreamAvatar~\cite{cao2024dreamavatar} being among the first works in this area that incorporate score distillation to optimize the human neural radiance field (NeRF) model. 
They utilize a deformable human NeRF model to render animatable scenes generated from diverse input texts.
TADA~\cite{liao2024tada} leverages SMPL-X~\cite{SMPL-X:2019} for modeling shape and UV texture, allowing for the rendering of more detailed 3D avatars. 
Recently, HumanNorm~\cite{huang2024humannorm} and Deceptive-Human~\cite{kao2023deceptive} have pushed the boundaries of 3D quality by incorporating additional 3D priors, including depth, normals, and pose information of human shapes.
HumanGaussian~\cite{liu2024humangaussian} successfully integrates Gaussian representations into the text-to-3D human task by training Gaussian splats with score distillation in a stable manner. However, it lacks animation capabilities, as it is designed exclusively for training static poses.

\section{Preliminaries}

\subsection{3D Gaussian Splatting}

3D Gaussian Splatting~\cite{kerbl20233d} is a powerful 3D modeling technique that enables real-time rendering by representing objects or scenes as collections of Gaussian splats.
Each splat $\mathcal{G}$ is characterized by its position $\mathbf{x}$, opacity $\alpha$, color $c$, and covariance matrix $\Sigma$, which defines its shape and spread through a scaling matrix $\boldsymbol{S}$ and rotation matrix $\boldsymbol{R}$. 
The overall image can be rendered by projecting each 3D Gaussian splat $\mathcal{G}$ onto the image plane. The pixel color is determined by accumulating the alpha values of the Gaussian splats as follows:

\begin{equation}
 C=\sum_i\left(\alpha_i^{\prime} \prod_{j=1}^{i-1}\left(1-\alpha_j^{\prime}\right)\right) c_i,
\end{equation}

\noindent where $\alpha_i^{\prime}$ represents the visibility $\alpha_i$ of the $i$-th splat, weighted by the probability density of $i$-th projected 2D Gaussian at the target pixel, and $c$ denotes the color value computed from spherical harmonics coefficients. 

It is mostly known for its real-time rendering while maintaining image quality compared to implicit neural representations.
While the original 3D Gaussians are optimized using a photometric loss based on the provided ground-truth pixels, our proposed method learns from the distillation loss derived from the diffusion model.


\subsection{Score Distillation}

Numerous powerful text-to-image diffusion models~\cite{rombach2022high, saharia2022photorealistic, ramesh2022hierarchical} have been proposed, demonstrating unprecedented image quality achieved through training on billions of text-image pairs.
Building on these diffusion models, Score Distillation Sampling (SDS) was introduced by DreamFusion~\cite{poole2022dreamfusion}, which distills prior knowledge from pre-trained 2D diffusion models to optimize the target 3D model.
When provided with a pre-trained diffusion model $\epsilon_\phi$, SDS optimizes the parameters of the 3D model $\theta$ using the gradient of the loss, which is represented as:

\begin{equation}
\nabla_\theta \mathcal{L}_{\mathrm{SDS}}=\mathbb{E}_{t, \epsilon, y}\left[w(t)\left(\epsilon_\phi\left(\mathbf{x}_t ; y, t\right)-\epsilon\right) \frac{\partial \mathbf{x}}{\partial \theta}\right],
\label{eq:sds}
\end{equation}

\noindent where $\mathbf{x}$ denotes the image rendered by the 3D model, $\mathbf{x}_t$ represents the rendered image with Gaussian noise $\epsilon$ added, and $y$ is the textual input encoded by the text encoder.
At each training iteration, different values of $t$ are sampled steering the rendered images closer to the distribution of real images.
While SDS produces successful distillation results by generating diverse 3D renderings that align with the given text inputs, it faces an over-saturation problem that critically impacts the creation of realistic details in 3D objects.



\section{Proposed Method}

An overview of our proposed method is described in \Figref{fig:framework}. 
We introduce \ourmodel, a human rendering model that generates text-aligned 3D human avatars from input textual descriptions and creates fully animatable scenes with random motion sequences by optimizing the deformable 3D Gaussian points using score distillation.

\subsection{Pose Deformable 3DGS}



To create animatable human avatars using 3D Gaussian Splatting, Gaussian points are defined in the canonical space and shared across all different poses. 
Starting from random locations on the canonical SMPL mesh surfaces as initialization, these Gaussian points are optimized to learn both the appearance consistent with the provided text and the ability to articulate various poses as the input pose varies.
To model different poses, each Gaussian splat is deformed according to the pose input using a rigid transformation method~\cite{qian20233dgs}.
Specifically, the Linear Blend Skinning (LBS) function is applied to transform 3D Gaussian splats from the canonical space to the observation space, where the target pose is specified as input.
As the human skeleton consist of $B$ joints, the transformation $\mathbf{T}$ is represented as the weighted sum of rigid bone transformations as: 


\begin{equation}
\mathbf{x}_o=\mathbf{T} \mathbf{x}_c = (\sum_{b=1}^B \mathbf{w}_b(\mathbf{x_c}) \mathbf{B}_b) \mathbf{x}_c,
\label{eq:transform}
\end{equation}

\noindent where $\mathbf{x}_c$ represents the position of Gaussian splats defined in the canonical space, and $\mathbf{x}_o$ denotes the position in the observation space.
$\mathbf{w}_b$ represents the blend weight for the $b$-th bone in the canonical space, which is further optimized.

$\mathbf{B}_b\in SE(3)$ is the transformation matrix of $b$-th skeleton part, mapping the bone's coordinates from the canonical space to the observation space.
Along with the transformation of positions, the rotation of Gaussian splats is also adjusted according to the equation $\mathbf{R}_o=\mathbf{T}_{1: 3,1: 3} \mathbf{R}_c$.
Note that the rigid bone transformation $\mathbf{B}$ can be computed from the given body pose (see Supplementary Material for details).

In 3DGS-avatar~\cite{qian20233dgs}, the blend weight $\mathbf{w}_b$ is optimized from scratch, as the ground truth provides clear guidance for determining the optimal values. 
In our method, on the other hand, the deformation must also be learned through distillation, resulting in weak guidance for optimizing the blend weights. 
Therefore, we propose to learn the residual blend weight relative to the SMPL blend weight $\mathbf{w}^{\mathbf{S}}$ as:

\begin{equation}
\mathbf{w}_b(\mathbf{x}_c)=\operatorname{norm}\left(f_{\theta_r}\left(\mathbf{x}_c\right)_b+\mathbf{w}^{\mathbf{S}}_{b}(\mathbf{x}_c)\right),
\label{eq:blend_weight}
\end{equation}

\noindent where $f_{\theta_r}$ is the MLP network which take the position coordinates of Gaussian splats as input to output residual blend weight values. 
The initial blend weight value $\mathbf{w}^{\mathbf{S}}$ can be computed based on the nearest vertex on the SMPL mesh by calculating the distance from the position coordinates of the Gaussian splats to each vertex.
The residual blend weight values are regularized by minimizing the $l2$-difference $\mathcal{L}_{\text{skinning}}$ between the predicted blend weight $\mathbf{w}_b$ and the initial SMPL blend weight $\mathbf{w}^{\mathbf{S}}$ across all positions of the Gaussian splats $\mathbf{x}_c$. 
By adopting this approach, we ensure that the pose transformation of SMPL is preserved from the initial stage, which effectively aids in converging to the appropriate blend weight.

\begin{figure*}[t]
\centering
\begin{tabular}{@{}c}
\includegraphics[width=1\linewidth]{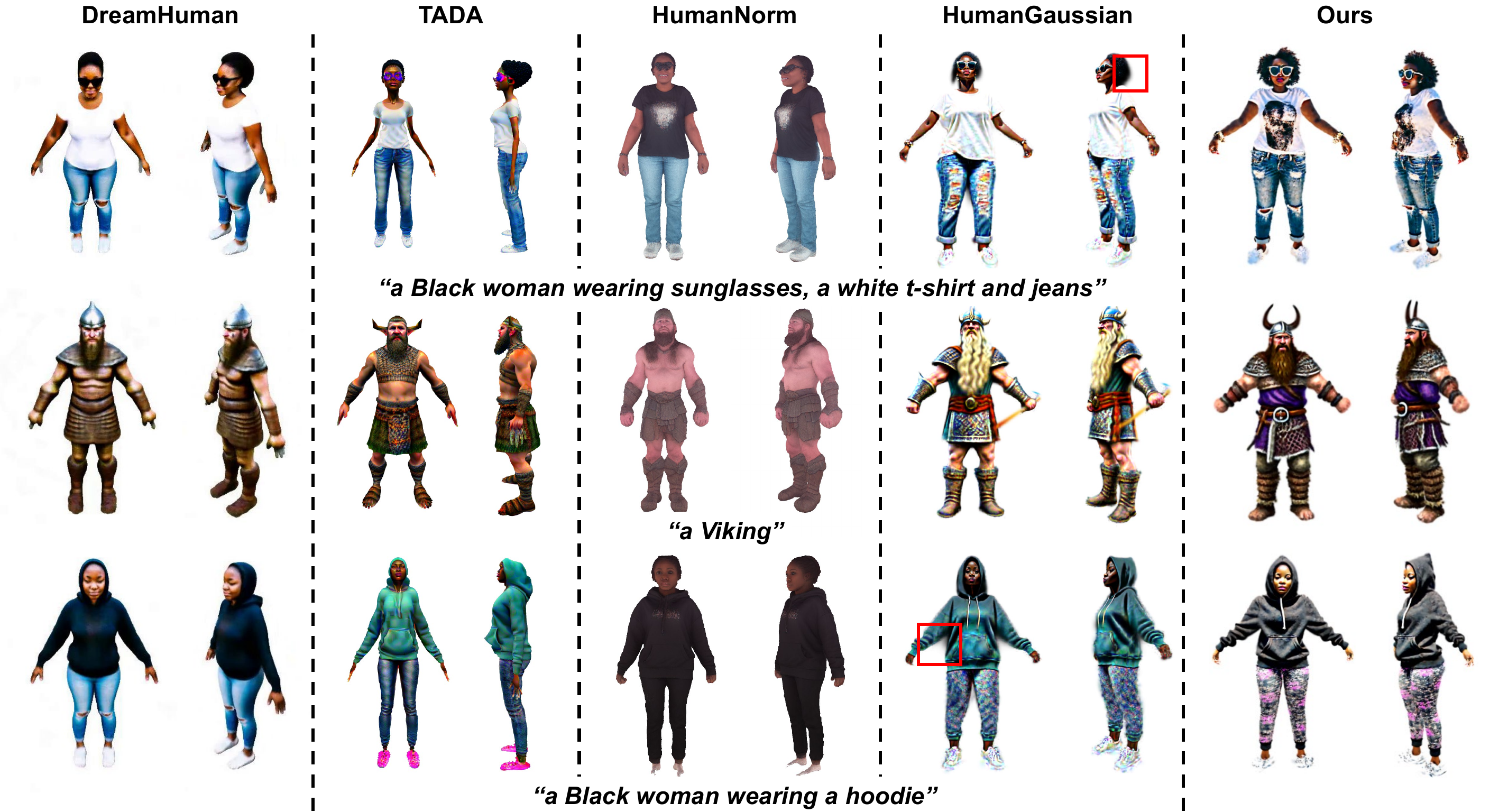} \\
\end{tabular}
\caption{{\bf Qualitative comparison of 3D human models in a static A-pose.} We evaluate our approach against recent state-of-the-art baselines using different prompts. For each method, two images are rendered from frontal and side views, respectively.
}
\label{fig:static}
\end{figure*}

\subsection{Dynamic Pose Guidance}
Previous methods~\cite{huang2024humannorm, liu2024humangaussian} have limited pose control, as they are primarily trained on static poses. 
To guarantee natural animatable scenes across diverse motion sequences, randomly posed images must be learned during the optimization of our 3D human model.
At each training step, we randomly sample poses from a normal distribution, where the mean pose is represented by a star pose.
By observing the randomly posed images during training, the deformable 3D human model learns a wide range of poses distilled from a pre-trained diffusion model in an end-to-end manner.

To provide robust pose-aware guidance when the complex pose images are rendered, we utilize ControlNet~\cite{zhang2023adding} which takes a pose image $p$ to generate pose-consistent images. 
As shown in \Figref{fig:framework}, the random pose sampled in each training iteration not only transforms the 3D Gaussian splats but is also rendered as pose images to be input into ControlNet, which adds spatial conditioning controls to the pre-trained diffusion models.
This integration facilitates more coherent distillation, ensuring that the generated outputs align with the sampled poses.

\begin{figure*}[t]
\centering
\def\arraystretch{0.2}
\begin{tabular}{@{}c}
\includegraphics[width=1\linewidth]{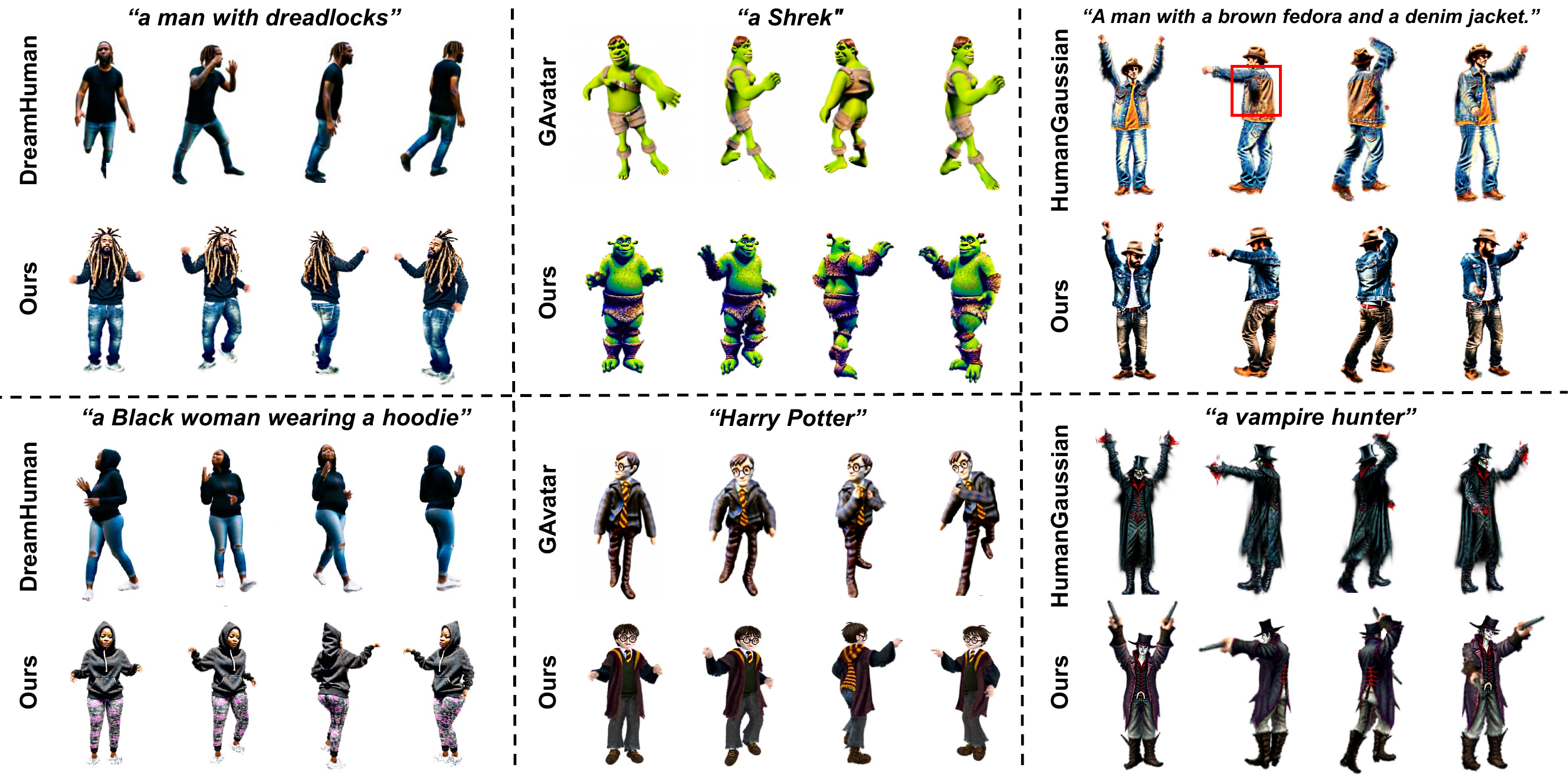} \\
\end{tabular}
\caption{{\bf Qualitative comparison of 3D human models in animated scenes.} We evaluate our approach against recent state-of-the-art baselines in a one-to-one manner. For each method, four images are rendered in different poses corresponding to each text prompt.}
\label{fig:animation}
\end{figure*}

\subsection{Adaptive Score Distillation}

By integrating pose guidance into score distillation, the score function in the gradient of the loss~\eqnref{eq:sds} is reformulated as $\epsilon_\phi\left(\mathbf{x}_t; y, t, p\right)$, where $p$ denotes the pose image conditioned on the diffusion model.
We then apply classifier-free guidance (CFG)~\cite{ho2022classifier} to score function and decompose the score matching difference into two components: the denoising score $\delta_n$ and the classifier score $\delta_c$, defined as:


\begin{equation}
\resizebox{.9\hsize}{!}{$
\begin{split}
\delta &= \delta_n + s \cdot \delta_c \\
       &= \left[\epsilon_\phi\left(\mathbf{x}_t ;t, p\right)-\boldsymbol{\epsilon}\right] 
       + s \cdot \left[\epsilon_\phi\left(\mathbf{x}_t ; y, t, p\right)-\epsilon_\phi\left(\mathbf{x}_t ; t, p\right)\right],
\end{split}
$}
\end{equation}
\noindent where $s$ is the guidance scale for CFG sampling.

While $\delta_c$ aims to direct the model towards high-density regions of real images conditioned on $y$, the denoising score $\delta_n$ often introduces excessive noise, resulting in blurry outputs due to averaging effects~\cite{katzir2023noise}. 
Previous methods~\cite{yu2023text, katzir2023noise} attempted to mitigate this by either removing the denoising score entirely or using a negative classifier score~\cite{katzir2023noise, liu2024humangaussian}. 
However, we empirically found that completely omitting the denoising score can produce artifacts, such as noise or shadow-like distortions, and that the generated outputs may deviate from the intended text description due to the high uncertainty of the classifier score (see \figref{fig:abl_ads}).

Based on the observation~\cite{katzir2023noise} that the denoising score  $\delta_n$ becomes overly noisy at smaller timesteps, while contributing to smoothness at larger timesteps,
we propose a simple yet effective technique called Adaptive Score Distillation (ASD). 
In this approach, the denoising score is selectively removed for timesteps below a threshold $\tau$ to mitigate noise, while being reintroduced for timesteps beyond $\tau$.
The score matching difference in ASD is defined as follows:



\begin{equation}
\delta = \delta_c \cdot \mathbb{1}_{t < \tau} + (\delta_n + s \cdot \delta_c) \cdot \mathbb{1}_{t \geq \tau},
\label{eq:ads}
\end{equation}

\noindent where $\tau$ can be adaptively defined to balance realistic details and smoothness. 

This adaptive approach allows our model to leverage the strengths of both score components: ensuring smoother high-level semantic alignment through the denoising score at higher timesteps, while maintaining sharp, detailed features by relying solely on the classifier score at lower timesteps. This ensures both output quality and clarity, effectively minimizing undesired noise in the 3D output.

\subsection{Training Objective}

\noindent {\bf Scale Regularization} 
The proposed adaptive score distillation successfully generates highly detailed objects; however, we observed some blurry results in certain samples, which were caused by Gaussian splats with large scales.
The blurriness observed around the surface arises from the SDS-based supervision, which is significantly more stochastic than photometric loss, as also noted in HumanGaussian~\cite{liu2024humangaussian}.
To overcome this problem, we propose to apply scale regularization loss during the optimization. 
Typically, the scales of Gaussian splats are adjusted through pruning techniques in the adaptive density control process, which involves removing Gaussian points that exceed a specified scale.
However, this often results in excessive pruning of Gaussian points, leading to a decrease in resolution and challenges in maintaining a balance with densification.
Additionally, points may be pruned at the boundaries, which can lead to a collapse of the human shape as training progresses. We found that imposing a regularization loss to limit the scale size yields the best output quality, preserving both the resolution of the Gaussian points and clear boundaries.

The scale regularization loss is defined as follow: 
\begin{equation}
\mathcal{L}_{\text {scale}}=\frac{1}{|\mathcal{P}|} \sum_{p \in \mathcal{P}} \max \left\{\max \left(\boldsymbol{S}_p\right), r\right\}-r
\label{eq:scale}
\end{equation}
\noindent where $\boldsymbol{S}_p$ represents the scalings of the 3D Gaussians. This loss regularizes the size of the 3D Gaussian points to ensure they do not exceed $r$.



\noindent {\bf Total Training Objective} Along with the proposed Adaptive Score Distillation $\mathcal{L}_{\text{ASD}}$, our total training objective functions for our method are written as: 
\begin{equation}
\begin{aligned}
&\mathcal{L}_{\text{total}}=\mathcal{L}_{\text{ASD}}+\lambda_{\text{scale}}\mathcal{L}_{\text{scale}} +\lambda_{\text{skinning}}\mathcal{L}_{\text{skinning}},
\end{aligned}
\end{equation}
\noindent where $\lambda_{\text{scale}}$ and $\lambda_{\text{skinning}}$ are hyperparameters determining the importance of each loss.

\section{Experimental Results}

\subsection{Qualitative Evaluation}

We compare our method with recent state-of-the-art approaches in two settings: static poses and animations generated from motion sequences.
For the static pose setting, we compare our method with DreamHuman~\cite{kolotouros2024dreamhuman}, TADA~\cite{liao2024tada}, HumanNorm~\cite{huang2024humannorm}, and HumanGaussian~\cite{liu2024humangaussian}, which are based on neural~\cite{mildenhall2021nerf}, mesh~\cite{shen2021deep}, and Gaussian representations~\cite{kerbl20233d}, respectively. For the animation results, we compare our method with DreamHuman, GAvatar~\cite{yuan2024gavatar}, and HumanGaussian.
Note that the official code of DreamHuman and GAvatar are not released, we use the results from their project pages.

As shown in \Figref{fig:static}, our method generates the most realistic 3D human results in terms of fine geometry and texture details when compared to the other baselines.
DreamHuman suffers from overly smoothed results, as it is represented with solid colors only, lacking details such as wrinkles on the clothing.
TADA produces more detailed textures than DreamHuman, but its human shapes appear unrealistic.
In the case of HumanNorm, areas like the face are generated with more detail through their refinement process, but the overall quality of the full body remains too simplistic, similar to DreamHuman.
HumanGaussian delivers the most plausible performance overall among comparison baselines, but it struggles to capture fine details like hair (see row 1), and leaves behind blurry artifacts between the arms (see row 3).
In contrast, our method generates high-frequency details and delivers clean results without any blurry artifacts. Moreover, as we will show shortly, the difference becomes even clearer in animation results.

We present the animation results in \Figref{fig:animation}. 
For each example, images with 4 different poses are visualized for each method.
DreamHuman leverages the structure of deformable human NeRF during training, enabling it to exhibit natural pose transformations. However, it still lacks realistic details, resulting in overly smooth and simplistic texture quality.
GAvatar also produces natural animation results by handling multiple poses through a primitive-based transformation. However, it struggles to capture complex geometric details, such as hair or loose clothing, which is a persistent issue associated with mesh representations.
HumanGaussian shows critical weaknesses in animation, as it is not designed to be animatable. 
While it provides a heuristic animation function by mapping Gaussian points to the SMPL-X~\cite{SMPL-X:2019} body mesh, 
it suffers from severe artifacts during pose changes (e.g., around the arms) due to mapping errors.
Additionally, it exhibits artifacts in occluded areas in the static pose, such as under the arms (see the 3rd example).
In contrast, our method produces realistic results in both geometry and texture for novel poses by capturing fine-grained details.

\begin{table}[t]
\centering
\renewcommand{\tabcolsep}{2.2mm}
\renewcommand{\arraystretch}{1.4}
\resizebox{0.9\linewidth}{!}{
\begin{tabular}{lccc}
\bottomrule
\bf Methods & \bf CLIP Score ↑ & \bf FID ↓ & \bf HPSv2 ↑\\ 
\toprule
HumanNorm~\cite{huang2024humannorm} & 27.81 & 5.41 & 0.256 \\
HumanGaussian~\cite{liu2024humangaussian} & 28.45 & 4.18 & 0.263\\ \hline
\rowcolor{cellcol} 
\bf GaussianMotion (ours) & \textbf{29.26} & \textbf{4.05} & \textbf{0.268} \\ 
\toprule
\end{tabular}
}
\caption{{\bf Quantitative comparisons with state-of-the-art text-to-3D human methods.} We evaluate CLIP score, FID, and HPS scores on rendered images.}
\label{tb:quanti}
\end{table}

\begin{figure}[t]
\centering
\def\arraystretch{0.2}
\begin{tabular}{@{}c}
\includegraphics[width=1\linewidth]{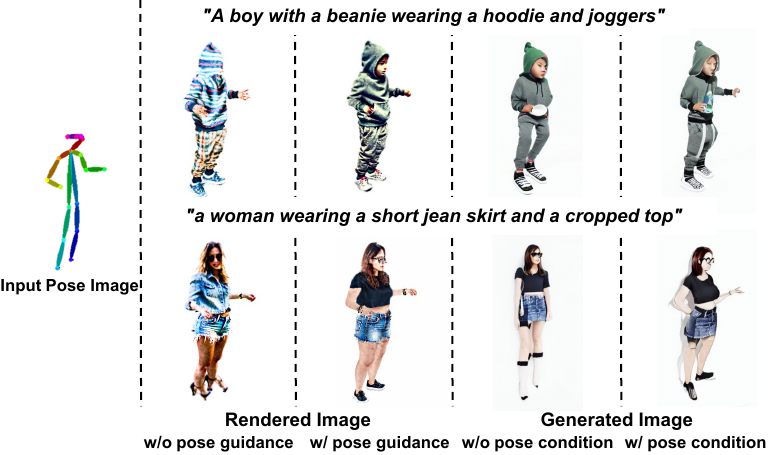} \\
\end{tabular}
\caption{{\bf Ablation studies on pose guidance.} 
We present rendered images from 3D models trained with and without pose guidance, with the input pose shown in the first column. Additionally, we show generated images sampled from noised rendered images, with and without pose conditioning. }
\label{fig:abl_pose}
\end{figure}

\subsection{Quantitative Evaluation}
Following HumanNorm~\cite{huang2024humannorm} and HumanGaussian~\cite{liu2024humangaussian}, we conducted a quantitative evaluation to assess the quality of the 3D rendered images produced by our method. 
We selected 30 text prompts from the list provided by HumanGaussian (see Supplementary Material for details) to constitute our test set.
First, we measured the Fréchet Inception Distance (FID)\cite{heusel2017gans}, which quantifies the similarity between feature distributions of real and generated images. We sampled 10 images per prompt using Stable Diffusion V1.5\cite{rombach2022high} for the real images, and used multiview images rendered from 10 azimuth angles in the range of $[-180^\circ, 180^\circ]$ for the generated image set.
Second, we evaluated the CLIP~\cite{hessel2021clipscore} and HPSv2~\cite{wu2023human} scores on the frontal views, which measure the similarity between embeddings encoded from the rendered images and the corresponding text. 
As detailed in \tabref{tb:quanti}, our method achieves the best score across all metrics, demonstrating that our method exhibits the best visual quality and consistency with the input text.

\noindent {\bf User Study} 
We conducted a user study to compare our method with recent state-of-the-art approaches~\cite{huang2024humannorm, liu2024humangaussian}. In this study, we presented pairs of multiview and animated scenes rendered by our method and one of the comparison methods. Using the same set of text prompts as in the quantitative evaluation, we created 30 A$\vs$B pairs and collected responses from 17 participants. Users were asked to assess three criteria: 1) Geometric Quality, 2) Texture Quality, and 3) Text Alignment. The results, summarized in \tabref{tb:user}, indicate that our method consistently outperforms the comparison methods across all criteria.

\begin{table}[t]
\centering
\renewcommand{\tabcolsep}{0.6mm}
\renewcommand{\arraystretch}{1.4}


\resizebox{0.999\linewidth}{!}{
\begin{tabular}{lccc}
\bottomrule
\bf Competitors & \bf Geometry & \bf Texture & \bf Text Consistency \\ \toprule
\vs. HumanNorm~\cite{huang2024humannorm} & 65.78 & 65.49 & 72.65 \\ 
\vs. HumanGaussian~\cite{liu2024humangaussian} & 84.51 & 89.31 & 67.45 \\ \toprule
\end{tabular}
}
\caption{{\bf User study with state-of-the-art text-to-3D human methods.} We present the preference percentage of our method compared to state-of-the-art methods.}
\label{tb:user}
\end{table}

\begin{figure}[t]
\centering
\def\arraystretch{0.2}
\begin{tabular}{@{}c}
\includegraphics[width=1\linewidth]{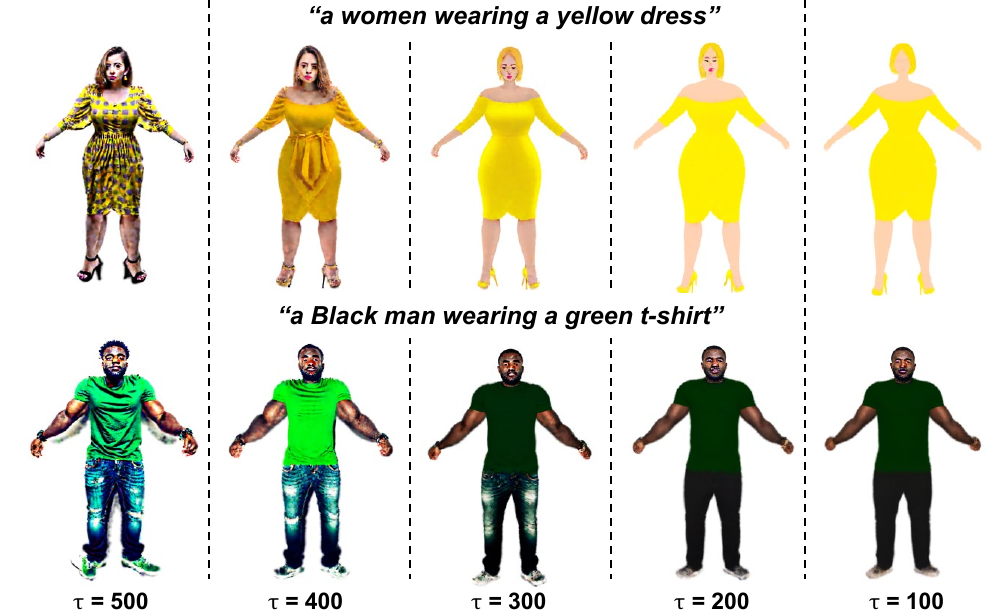} \\
\end{tabular}
\caption{{\bf Ablation studies on $\tau$ changes in Adaptive Score Distillation.} Adjusting the $\tau$ value achieves a balanced result, preserving fine details while minimizing noise.}
\label{fig:abl_ads}
\end{figure}

\subsection{Ablation Studies}
\noindent {\bf Pose Guidance}
We demonstrate the impact of incorporating pose guidance, which significantly enhances our method's performance, by comparing it with a setting where the model is trained without pose guidance (using only text input for the diffusion model). As shown in \figref{fig:abl_pose}, when trained without pose guidance, the model fails to capture the correct pose and orientation (for example, no face is generated in row 1). 
In the right part of the figure, we also show a generated image sampled from a rendered image with Gaussian noise added at $t=600$.  
The image generated without pose conditioning does not accurately reflect the input pose shown in the reference, implying that the diffusion model fails to produce pose-consistent scores during distillation, significantly hindering our 3D model from learning the correct poses.

\noindent {\bf Adaptive Score Distillation} 
Here, we demonstrate the effectiveness of Adaptive Score Distillation and show how the generation results vary with different $\tau$ values in \eqnref{eq:ads}. First, we follow the annealed distillation time schedule from~\cite{wang2024prolificdreamer}, reducing the maximum distillation timestep to 500 from the middle of the training process, which improves visual quality in the later stages.
We then decrease $\tau$ from 500 (corresponding to training with only the classifier score) in intervals of 100.
As shown in \Figref{fig:abl_ads}, the smoothness derived from the denoising score and the low-level details from the classifier score are interpolated as $\tau$ changes. 
With the highest $\tau$ value, our method fails to produce clean results, incorporating undesired noise such as floating artifacts and shadows. Additionally, we observe that some samples deviate significantly from the original text description (see row 1).
On the other hand, with lower value of $\tau$, which approach the naive SDS as it decrease, it shows results that are oversaturated and far from realistic.
In contrast, by adjusting the $\tau$ value (around 400), we were able to achieve the most balanced results, satisfying both fine details and a lack of noise.

\noindent {\bf Scale Regularization} 
We empirically found that scale regularization significantly improves generation quality by reducing blurriness and enhancing fine details through small Gaussian points. In our baseline setting (a), Gaussian points are pruned if their scaling factor exceeds 0.1. To evaluate the effect of scale regularization versus pruning, we also tested settings (b) and (c), where points with scaling factors above 0.02 and 0.01 are pruned, respectively. Lastly, setting (d) applies scale regularization without altering the pruning threshold.
While setting (a) suffers from blurriness that severely harms visual quality, settings (b) and (c) successfully remove it but at the cost of excessive pruning, which reduces resolution and causes loss of shape as training progresses. In contrast, as shown in (d), applying a regularization loss with $r$ set to 0.01 in \eqnref{eq:scale} yields the best visual quality, preserving  both the resolution of Gaussian points and clear boundaries.

\begin{figure}[t]
\centering
\def\arraystretch{0.2}
\begin{tabular}{@{}c}
\includegraphics[width=1\linewidth]{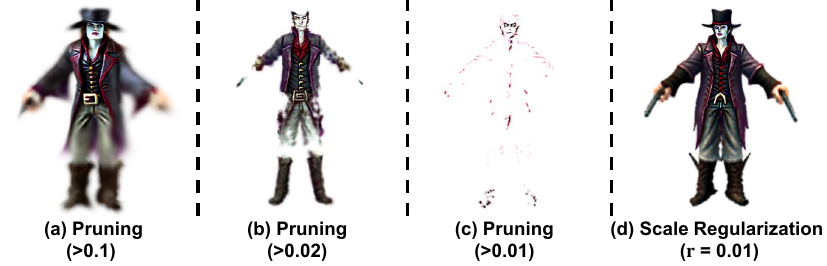} \\
\end{tabular}
\caption{{\bf Ablation studies on scale regularization against scaling-based pruning.} We present frontal views rendered using each trained 3D model, which are trained with scaling-based pruning and scale regularization, respectively.}
\label{fig:abl_scale}
\end{figure}

\section{Conclusion}


In this paper, we present \ourmodel, a novel approach for generating animatable 3D human models from text descriptions using Gaussian Splatting. Our method overcomes the limitations of existing approaches, including fidelity, efficiency, and dynamic pose control, by combining deformable Gaussian Splatting with pose-aware score distillation. By densely sampling random poses during training, our model learns diverse pose variations and fine details, resulting in high-quality renderings. The proposed Adaptive Score Distillation further refines output quality, balancing detail and smoothness. Experimental results demonstrate that our method outperforms state-of-the-art baselines, offering an efficient, detailed, and pose-flexible solution for creating 3D avatars from text.

{
    \small
    \bibliographystyle{ieeenat_fullname}
    \bibliography{main}
}


\end{document}